\documentclass[11pt,a4paper]{article}
\usepackage[hyperref]{ranlp2021}
\usepackage{times}
\usepackage{latexsym}

\usepackage{graphicx}
\usepackage{listings}
\usepackage{lipsum}
\usepackage{amssymb}

\usepackage{array}
\newcolumntype{P}[1]{>{\centering\arraybackslash}p{#1}}

\newcommand{\comment}[1]{}

\usepackage{microtype}

\aclfinalcopy % Uncomment this line for the final submission
\def\aclpaperid{40} %  Enter the acl Paper ID here

\title{PyEuroVoc: A Tool for Multilingual Legal Document Classification with EuroVoc Descriptors}

\author{Andrei-Marius Avram, Vasile Păiș, Dan Tufiș \\
  Research Institute for Artificial Intelligence “Mihai Drăgănescu” \\
  Bucharest, Romania \\
  \texttt{\{andrei.avram,vasile,tufis\}@racai.ro} }

\date{}

\begin{document}
\maketitle
\begin{abstract}
EuroVoc is a multilingual thesaurus that was built for organizing the legislative documentary of the European Union institutions. It contains thousands of categories at different levels of specificity and its descriptors are targeted by legal texts in almost thirty languages. In this work we propose a unified framework for EuroVoc classification on 22 languages by fine-tuning modern Transformer-based pretrained language models. We study extensively the performance of our trained models and show that they significantly improve the results obtained by a similar tool - JEX - on the same dataset. The code and the fine-tuned models were open sourced, together with a programmatic interface that eases the process of loading the weights of a trained model and of classifying a new document.
\end{abstract}

\section{Introduction}

EuroVoc\footnote{\url{https://data.europa.eu/data/datasets/eurovoc}} is a multilingual thesaurus which was originally built up specifically for processing the documentary information of the EU institutions. The covered fields are encompassing both European Union and national points of view, with a certain emphasis on parliamentary activities. The current release 4.4 of EuroVoc was published in December 2012 and includes 6,883 IDs for thesaurus concepts (corresponding to the preferred terms), classified into 21 domains (top-level domains), further refined into 127 subdomains. Additional forms of the preferred terms are also available and are assigned the same ID, subdomains and top-level domains.

Multilingual EuroVoc thesaurus descriptors are used by a large number of European Parliaments and Documentation Centres to index their large document collections. The assigned descriptors are then used to search and retrieve documents in the collection and to summarise the document contents for the users. As EuroVoc descriptors exist in one-to-one translations in almost thirty languages, they can be displayed in a language other than the text language and give users cross-lingual access to the information contained in each document.

One of the most successful recent approaches in document and text classification involves fine-tuning large pretrained language models on a specific task \cite{adhikari2019docbert,nikolov2019nikolov}. Thus, in this work we propose a tool for classifying legal documents with EuroVoc descriptors that uses various flavours of Bidirectional Encoder from Transformers (BERT) \cite{devlin2019bert}, specific to each language. We evaluated the performance of our models for each individual language and show that our models obtain a significant improvement over a similar tool - JEX \cite{steinberger2012jrc}. The The models were further integrated into the RELATE platform \cite{pais2020} and an API was provided through the PythonPackage Index (PyPi) interface\footnote{\url{https://pypi.org/project/pyeurovoc/}} that facilitates the classification of new documents. The code used to train and evaluate the models was also open-sourced\footnote{\url{https://github.com/racai-ai/pyeurovoc}}.

The rest of the paper is structured as follows. Section \ref{sec:background} presents other works in the direction of EuroVoc classification. Section \ref{sec:dataset} provides several statistics with regard to the corpus used to train and test the models in the tool. Section \ref{sec:models} presents the approach used in fine-tuning the pretrained language models and the exact BERT variants used for each language, together with a vocabulary statistics of the model's tokenizer on the legal dataset. Section \ref{sec:experiments} outlines our evaluation setup and the results of our experiments, while Section \ref{sec:prog_interf} presents the programmatic interface. Finally, the paper is concluded in Section \ref{sec:conclusion}.
 
\section{Related Work}
\label{sec:background}

JEX \cite{steinberger2012jrc} is a multi-label classification software developed by Joint Research Centre (JRC), that was trained to assign EuroVoc descriptors to documents that cover the activities of the EU. It was written entirely in Java and it comes with 4 scripts (both batch and bash) that allows a user to pre-process a set of documents, train a model, postprocess the results and evaluate a model. Each script is easily configurable from a properties file that contains most of the necessary parameters. The toolkit also comes with a graphical interface that allows a user to easily label a set of new documents (in plain text, XML or HTML) or to train a classifier on their own document collections. 

The algorithm used for classification was described in \cite{pouliquen2006automatic} and it consists in producing a list of lemma frequencies from normalized text, and their weights, that are statistically related to each descriptor, entitled in the paper as associates or as topic signatures. Then, to classify a new document, the algorithm picks the descriptors of the associates that are the most similar to the list of lemma frequencies of the new document. The initial release consisted of 22 pretrained classifiers, each corresponding to an official EU language.

Boella et al. \cite{boella2012multi}, while focusing on the Italian JRCAcquis-IT corpus, presents a technique for transforming multi-label data into mono-label that is able to maintain all the information as in \cite{tsoumakas2007multi}, allowing the use of approaches like Support Vector Machines (SVM) \cite{joachims1998text} for classification. Their proposed method allows an F1 score of 58.32 (an increase of almost 8\% compared to the JEX score of 50.61 for the Italian language). 

Šarić et al. \cite{vsaric2014multi} further explores SVM approaches for classification of Croatian legal documents and report an F1 score of 68.6. Unfortunately, this is not directly comparable with the JEX reported results since the training corpus is a different collection called NN13205. Furthermore, the categories being used for the gold annotation represent an extended version of EuroVoc for Croatian, called CroVoc.

Studies, such as in \cite{collobert2011natural}, have shown that neural word embeddings can store abundant semantic meanings and capture multi-aspect relations into a real-valued matrix, when trained on large unlabeled corpora using neural networks. Considering a vocabulary $V$, an embeddings representation can be learned by means of a neural network resulting into an association of a real-valued vector $W_n$ of size $n$ to each word. Two neural network methods for automatically learning distributed representation of words from a large text corpus can be con-sidered: Skip-gram and continuous bag of words (CBOW) \cite{DBLP:journals/corr/abs-1301-3781}. In the case of CBOW, a neural network is trained to predict the middle word given a context, while Skip-gram uses a single word as input and tries to predict past and future words. Bojanowski et al. \cite{bojanowski2017enriching} introduced a method for runtime representation for unknown words by means of averaging pre-trained character n-grams, also known as subword information.

BERT has also been used to classify legal documents with EuroVoc labels, with most of the work focusing on the English language. In \cite{chalkidis2019large}, the authors studied the problem of Large-Scale Multi-Label Text Classification (LMTC) for few- and zero-shot learning and released a new dataset composed of 57k samples from EUROLEX on which several models were tested. The results showed that BERT obtained superior performance in all but zero-shot classification.

\section{Dataset Statistics}
\label{sec:dataset}

\begin{figure*}
    \centering
    \includegraphics[width=\textwidth]{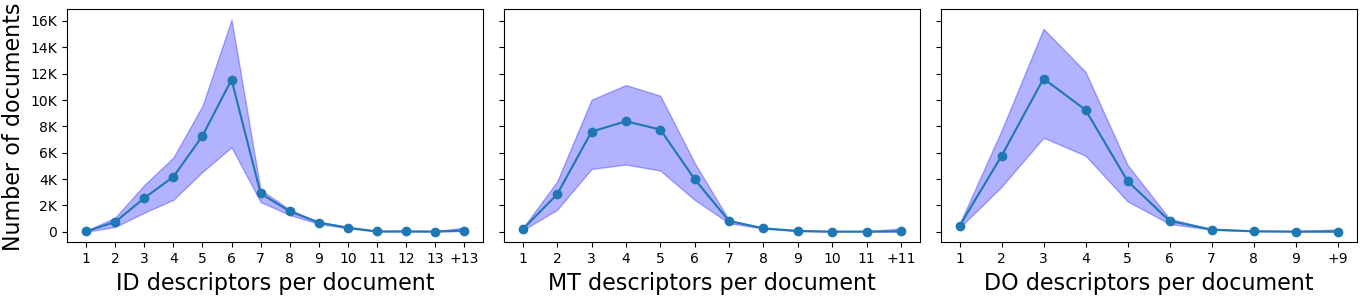}
    \caption{Distribution of the mean ID, MT and DO descriptors per document with fill between the minimum and the maximum values.}
    \label{fig:desc_distrib}
\end{figure*}

\begin{figure*}
    \centering
    \includegraphics[width=\textwidth]{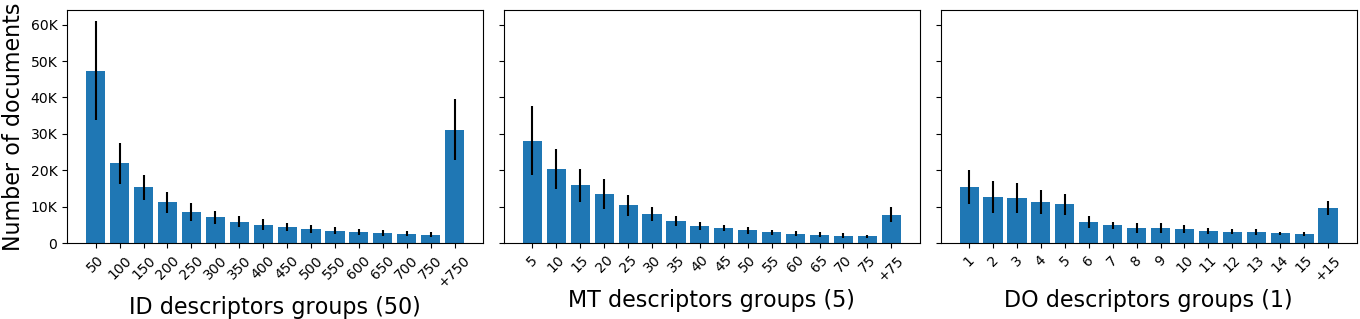}
    \caption{Number of documents that contains the most frequent ID, MT and DO descriptor groups. Each ID group contains 50 descriptors, each MT group contains 5 descriptors and each DO group contains only 1 descriptor. The standard deviation between languages is also displayed on the y axis.}
    \label{fig:hist_distrib}
\end{figure*}

The training of BERT models for the 22 languages was done using the same dataset that was used for training the JEX models. The dataset is composed of two parallel corpora from the legal domain, JRC-Acquis \cite{steinberger2006jrc} and the Publications Office of the European Union (OPOCE), that were manually labeled with over 6,700 EuroVoc descriptors identifiers (ID). The EuroVoc descriptors are hierarchically organised and can be converted into higher level Microthesaurus labels (MT) and further into top-level domains (DO). 

The number of documents in the dataset range from 17,858 documents in Maltese to 41,989 documents in French. Each document is labeled with multiple ID descriptors, having an average of 6 ID descriptors which can be equivalently converted to 5 MT descriptors or 4 DO descriptors. In Figure \ref{fig:desc_distrib} we depict the distribution of the average number of ID, MT and DO descriptors per document, together with the difference between the minimum and the maximum number of documents per descriptor. 

The ID, MT and DO descriptors distributions are also highly unbalanced. Figure \ref{fig:hist_distrib} depicts the number of documents of the most frequent ID, MT and DO descriptors, organised in groups of 50, 5 and 1, respectively. Each group contains the sum of the number of documents that are labeled with each descriptor in the respective group. As it can be observed, in each subplot, the number of documents that contain the descriptors from the first few groups is higher than the number of document that contain all the other descriptors.

\section{Methodology}
\label{sec:models}

The proposed approach for classifying the legal documents found in the two corpora is to fine-tune a pre-trained BERT on each of the 22 languages. We follow the method introduced in \cite{devlin2019bert} where a simple feed-forward network with the weights $W \in \mathbb{R}^{E \times M}$, $E$ is embedding size of BERT and $M$ is the number of classes, and bias $b \in \mathbb{R}^M$ is put on top of the embedding of the first token $C$ (\texttt{[CLS]}) to create the output logits of the classification problem for the ID descriptors\footnote{The MT and DO descriptors are predicted by using a direct mapping scheme}. The sigmoid $\sigma$ function is then applied to produce independent probability distributions $\hat{y}$ over each class:

\begin{equation}
    \hat{y} = P(y | x) = \sigma (C^T W + b)
\end{equation}

Additionally, a dropout of 0.1 is applied on the feed-forward layer to regularize the model.

The models are optimized by reducing the loss $\mathcal{L}$ computed as the average binary cross-entropy between the output probabilities $\hat{y}$ and the target classes $y$, over the $M$ classes (ID descriptors).:

\begin{equation}
    \mathcal{L}  = - \frac{1}{M} \sum_{i=1}^M y_i \log \hat{y_i} + (1 - y_i) \log (1 - \hat{y_i})
\end{equation}

Because the flavours of BERTs vary from one language to another, the choice of the initial models for each language was made by using the following heuristic, based on the corpora used for pretraining: \textit{Legal $>$ Monolingual (Mono) $>$ Wikipedia (Wiki) $>$ Multilingual (Multi)}. The heuristic is experimentally supported by \cite{chalkidis2020legal} that showed that language models obtain superior performance on the legal domain when they are pretrained on legal corpora and by \cite{pyysalo2020wikibert} that outlined the superiority of BERTs pretrained on monolingual Wikipedia over multilingual BERT (mBERT). Also, it was empirically proven that the performance of the language models improves when they are pretrained on larger corpora \cite{liu2019roberta} and for this reason we expect most of the general monolingual models to obtain better result than Wikipedia BERTs. Thus, given the existing open-sourced models for each language, we use the following taxonomy in our experiments: (Figure \ref{fig:models_tax}) \footnote{To the best of our knowledge, not all languages have publications for their monolingual versions of BERT, so we attached a corresponding URL in these cases.}:

\begin{figure}
    \centering
    \includegraphics[width=0.43\textwidth]{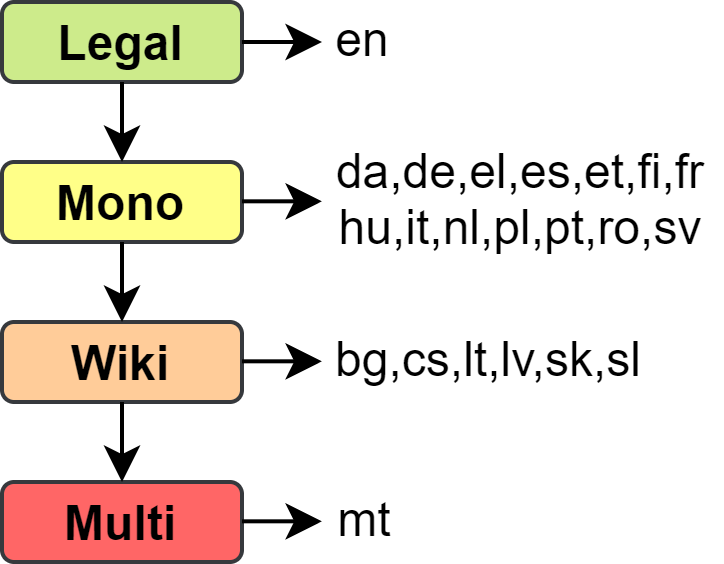}
    \caption{The proposed categories to divide the pre-trained language models and their corresponding languages.}
    \label{fig:models_tax}
\end{figure}

%which states that, due to performance reasons, monolingual legal pre-trained models are preferred over monolingual general pre-trained models \cite{chalkidis2020legal} which are in turn preferred over multilingual pre-trained models \cite{pires2019multilingual}. Moreover, we also choose Wiki BERTs over the multilingual ones, as they have been shown to obtain superior results for most of the languages that they were tested on\cite{pyysalo2020wikibert}.

\begin{itemize}
    \item \textbf{Legal}: English (en) - Legal BERT \cite{chalkidis2020legal}.
    
    \item \textbf{Mono}: Danish (da) - Danish BERT \footnote{\url{https://github.com/botxo/nordic_bert}}, German (de) - German BERT\footnote{\url{https://huggingface.co/bert-base-german-cased}}, Greek (el) - Greek BERT \cite{koutsikakis2020greek}, Spanish (es) - Spanish BERT \cite{canete2020spanish}, Estonian (et) - EstBERT \cite{tanvir2020estbert}, Finnish (fi) - Finnish BERT \cite{virtanen2019multilingual}, French (fr) - CamemBERT \cite{martin2020camembert}, Hungarian (hu) - huBERT \cite{erzsebetdavid}, Italian (it) - Italian BERT \footnote{\url{https://huggingface.co/dbmdz/bert-base-italian-cased}}, Dutch - BERTje \cite{de2019bertje}, Polish (pl) - PolBERT \cite{Kleczek2020}, Portuguese (pt) - BERTimabau\cite{souza2020bertimbau}, Romanian (ro) - Romanian BERT \cite{dumitrescu2020birth}, Swedish (sv) - Swedish BERT \cite{swedish-bert}.
    
    \item \textbf{Wiki}: Bulgarian (bg) - WikiBERT-BG, Czech (cs) - WikiBERT-CS, Lithuanian (lt) - WikiBERT-LT, Latvian (lv) - WikiBERT-LV, Slovak (sk) - WikiBERT-SK, Slovene (sl) - WikiBERT-SL.
    
    \item \textbf{Multi}: Maltese (mt) \footnote{\url{https://huggingface.co/bert-base-multilingual-cased}}.
\end{itemize}

\begin{table}
    \centering
    \begin{tabular}{lcc}
          & \textbf{Tokens/} & \textbf{UNK/} \\
          \textbf{Vocabulary} & \textbf{Word} & \textbf{Word}  \\
          \hline
            WikiBERT-BG (bg) & 1.64 & 5e-3 \\
            WikiBERT-CS (cs) & 2.15 & 1e-4  \\
            Danish BERT (da) & 1.51 & 6e-3  \\
            German BERT (de) & 1.64 & 1e-3 \\
            Greek BERT (el) & 1.44 & 8e-5 \\
            Legal BERT (en) & 1.28 & 3e-4  \\
            Spanish BERT (es) & \textbf{1.25} & 6e-3  \\
            EstBERT (et) & 1.87 & 2e-4 \\
            Finnish BERT (fi) & 1.72 & 1e-3 \\
            CamemBERT (fr) & 1.40 & \textbf{0}  \\
            huBERT (hu) & 1.80 & 2e-4  \\
            Italian BERT (it) & 1.36 & 2e-4 \\
            WikiBERT-LT (lt) & 2.05 & 1e-4 \\
            WikiBERT-LV (lv) & 2.15 & 5e-4  \\
            mBERT (mt) & 2.87 & 1e-2  \\
            BERTje (nl) & 1.42 & 1e-3 \\
            PolBERT (pl) & 1.52 & 6e-5 \\
            BERTimbau (pt) & 1.47 & 5e-3  \\
            Romanian BERT (ro) & 2.31 & 1e-4  \\
            WikiBERT-SK (sk) & 2.14 & 1e-4 \\
            WikiBERT-SL (sl) & 1.76 & 4e-4  \\
            Swedish BERT (sv) & 1.45 & 5e-4 \\
          \hline
    \end{tabular}
    \caption{Vocabulary statistics of each pretrained BERT model on the legal dataset.}
    \label{tab:vocab-stats}
\end{table}

The vocabulary of BERT plays an important role in the final performance of the model. Broadly speaking, the fewer tokens each word is split into, the better the language model is expected to perform. In Table \ref{tab:vocab-stats} we depict the average number of tokens per word and the average number of unknown (\texttt{UNK}) tokens per word on the legal dataset for each tokenizer of the 22 BERT models. As it can be observed, the lowest number of tokens per word is achieved by the Spanish BERT with 1.25 followed closely by the Legal BERT with 1.28. When looking at unknown tokens per word, CammemBERT tokenizer leads the leaderboard with no unknown words when tokenizing the dataset. On the other hand, the highest number of tokens and unknown tokens per word was achieved on Maltese due to use of mBERT instead of a monolingual model.

The legal documents in the corpus can be rather long and exceed the maximum number of tokens of 512 allowed by the BERT models. To mitigate this, we simply keep only the first 512 in the document and discard the rest. This method has been shown to lead to approximately the same performance as considering the whole document \cite{chalkidis2019large}.

\section{Experiments}
\label{sec:experiments}

\begin{table*}[ht]
    \centering
    \begin{tabular}{lP{1.3cm}P{1.3cm}P{1.3cm}P{1.3cm}P{1.3cm}P{1.3cm}}
          \multicolumn{1}{c}{} & \multicolumn{3}{c}{\textbf{JEX}} & \multicolumn{3}{c}{\textbf{BERT}} \\
          \hline
          \textbf{Language} & \textbf{ID} \textit{\textbf{F1@6}} & \textbf{MT} \textit{\textbf{F1@5}} & \textbf{DO} \textit{\textbf{F1@4}} & \textbf{ID} \textit{\textbf{F1@6}} & \textbf{MT} \textit{\textbf{F1@5}} & \textbf{DO} \textit{\textbf{F1@4}} \\
          \hline
          
          Bulgarian (bg) & 47.66 & 60.45 & 68.34 & 80.86 & 85.97 & 88.17 \\
          Czech (cs) & 47.81 & 60.03 & 68.62 & 80.10 & 84.18 & 87.36 \\
          Danish (da) & 49.79 & 62.45 & 71.24 & 79.56 & 80.82 & 87.19 \\
          German (de) & \textbf{50.65} & \textbf{63.15} & \textbf{72.23} & 79.69 & 85.26 & 89.90 \\
          Greek (el) & 49.65 & 61.34 & 70.41 & 71.19 & 78.69 & 80.06 \\
          English (en) & 50.15 & 62.86 & 71.60 & 78.95 & 82.47 & 87.95 \\
          Spanish (es) & 50.42 & 62.47 & 71.19 & 78.33 & 82.02 & 86.22 \\
          Estonian (et) & 48.98 & 61.19 & 69.57 & 75.90 & 81.53 & 84.19 \\
          Finnish (fi) & 49.41 & 61.65 & 70.76 & 78.07 & 82.48 & 87.60 \\
          French (fr) & 50.61 & 62.87 & 71.58 & 75.01 & 77.72 & 81.45 \\
          Hungarian (hu) & 49.71 & 62.22 & 70.34 & 73.01 & 78.50 & 83.11 \\
          Italian (it) & 49.53 & 61.15 & 70.14 & 77.82 & 80.59 & 87.43 \\
          Lithuanian (lt) & 48.26 & 61.20 & 70.31 & 76.49 & 78.26 & 83.92 \\
          Latvian (lv) & 47.79 & 59.14 & 68.18 & 80.07 & 84.98 & 87.22 \\
          Maltese (mt) & 44.36 & 55.99 & 66.11 & 69.06 & 72.15 & 81.46 \\
          Dutch (nl) & 50.64 & 62.40 & 71.00 & 81.33 & 83.97 & 87.11 \\
          Polish (pl) & 46.86 & 59.15 & 68.17 & 76.40 & 79.54 & 85.17 \\
          Portuguese (pt) & 50.41 & 62.55 & 71.57 & 82.92 & 86.61 & 91.28 \\
          Romanian (ro) & 47.13 & 60.18 & 69.39 & 80.90 & 86.12 & 88.40 \\
          Slovak (sk) & 46.34 & 58.36 & 67.30 & 83.40 & 82.29 & 85.25 \\
          Slovenian (sl) & 49.96 & 62.63 & 70.81 & \textbf{84.90} & \textbf{87.37} & \textbf{91.72} \\
          Sweedish (sv) & 50.32 & 62.21 & 71.04 & 77.88 & 81.79 & 84.24 \\
          
          \hline
    \end{tabular}

    \caption{Evaluation results of JEX and BERT for ID, MT and DO descriptors.}
    \label{tab:eval_results}
\end{table*}

\subsection{Evaluation Setup}

Because the original splits used for training and evaluating the JEX models were not made publicly available, we united the JRC-Acquis and OPOCE datasets for each language and split it 5 times in train, validation and test sets using different seeds. Moreover, in order to preserve the class balance across the sets in one split, we employed an iterative stratification splitting approach as proposed in \cite{sechidis2011stratification} and kept an approximate ratio of 80\% train, 10\% validation and 10\% test for fine-tuning and evaluating the pre-trained language models and a ratio of 90\% train and 10\% test for training and evaluating the JEX models.

The pre-trained language models were fine-tuned for 30 epochs, using a batch size of 8 and the AdamW optimizer \cite{loshchilov2018decoupled} whose learning rate was decayed by a linear scheduler peaking at 6e-5, in order to reduce the oscillations in the later stages of training due to the high values of the learning rate. We also clipped the gradients \cite{pascanu2013difficulty} whose norm had a value over 5 and used a learning rate warm-up over the first epoch to alienate the effects of forgetting the knowledge learned by Transformer models in the pre-training phase. The final weights of each fine-tuned language model were the ones that obtained the lowest loss on the validation set during training.

The training and evaluation of JEX models followed the approach described in \cite{steinberger2012jrc}. Both JEX models and the pre-trained language models were trained five times on each split with the results averaged over all test splits. We also used the validation splits for early stopping and fine-tune the hyperparameters of the BERT models.

\subsection{Evaluation Metrics}

\begin{figure*}
    \centering
    \includegraphics[width=\textwidth]{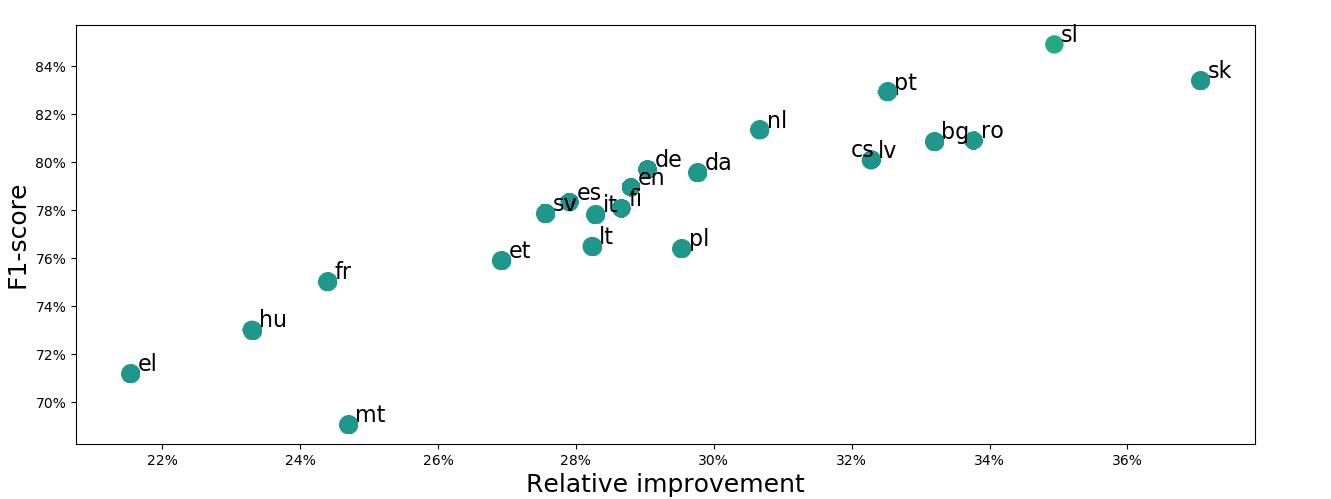}
    \caption{Performance of BERT models on ID multi-label classification relative to the performance of JEX models on the same language.}
    \label{fig:scatter_performance}
\end{figure*}

Most used metrics for evaluating LMTC models are are the precision (\textit{P@K}), the recall (\textit{R@K}) and their harmonic mean, known as F1 score (\textit{F1@K}), over the top $K$ predicted labels. These metrics usually unfairly penalize documents that have fewer or more labels than $K$, but we still use them because they allow a direct comparison with the original results of JEX. The three metrics are defined as follows:

\begin{equation}
    P@k = \frac{1}{k} \sum_{l \in r_k(\hat{y})}^k y_l
\end{equation}

\begin{equation}
    R@k = \frac{1}{n} \sum_{l \in r_k(\hat{y})}^k y_l
\end{equation}

\begin{equation}
    F1@k = 2 \cdot \frac{P@k \cdot R@k}{P@k + R@k}
\end{equation}

where $k$ is the number of labels to be used for comparison, $n$ is the number of true labels of the respective document, $y \in \{0,1\}^L$ is the vector of the true labels, $\hat{y} \in \mathbb{R}^L$ is the vector of predicted labels and $r_k(\hat{y})$ is a function that selects the index of the $k$th largest value in the prediction labels.
 
As the statistics in Section \ref{sec:dataset} have shown, the average number ,per document, of ID descriptors is 6, of MT descriptors is 5 and of DO descriptors is 4. Thus, we evaluate both JEX and BERT by using the F1 score for 6 labels on ID descriptors (\textit{F1@6}), for 5 labels on MT descriptors (\textit{F1@5}) and for 4 labels on DO descriptors (\textit{F1@4}).

\subsection{Results}

The results for both JEX and the BERT models on the 22 languages by using the cross-validated dataset are outlined in Table \ref{tab:eval_results}. The BERT models obtained a significant improvement over JEX on each language, ranging from an enhancement of 21.54\% (el), 14.85\% (fr) and 9.65\% (el) to an enhancement of 37.06\% (sk), 25.94\% (ro) and 19.83\% (sl) for ID, MT and DO descriptors, respectively. The highest F1 scores with JEX were achieved on German with 50.65\% \textit{F1@6}, 63.15\% \textit{F1@5}, 72.23\% \textit{F1@4}, and the highest F1 scores with the BERT models were achieved on Slovenian with 84.90\% \textit{F1@6}, 87.37\% \textit{F1@5}, 91.72\% \textit{F1@4} for ID, MT and DO descriptors, respectively. On the other hand, the lowest scores were obtained on Maltese. This might be due to the low number of documents compared to the other languages \cite{steinberger2012jrc}, but also because, in the case of the BERT variant, we use a multilingual model instead of a monolingual one.

Figure \ref{fig:scatter_performance} depicts the \textit{F1@6}-scores obtained by the BERT models on multi-label ID classification relative to the scores obtained by JEX models in the same language. One interesting aspect that can be observed in the plot is that although the mBERT used for Maltese obtained the lowest F1-score, its relative improvement over JEX is higher than of the other three languages that use monolingual models: Greek, Hungarian and French, mostly because the \textit{F1@6}-score obtained by JEX on Maltese is lower when compared to the other three and thus the difference would normally be larger.

The variance of scores between languages is higher for the BERT models\footnote{BERT high-low differences: ID - 15.84\%, MT - 15.22\%, DO - 10.26\%.} than for the JEX models\footnote{JEX high-low differences: ID - 6.29\%, MT - 7.16\%, DO - 6.12\%}. This happens because the BERT models were pretrained beforehand on various corpora taken from different sources and aspects like the quality of the corpus or the domain match greatly influenced the resulted fine-tuning performance. One interesting result is that WikiBERT obtained some of the highest scores and that Legal BERT also did not perform as well as expected, thus partially contradicting the heuristic introduced in the previous section. Due to time and resource constraints, we leave a detailed study of the heuristic for future work.

\subsection{Comparison with State-of-the-Art}

The state-of-the-art (SOTA) for EuroVoc multi-label classification was presented in \cite{chalkidis2019large} by using the original BERT-base. The model was trained and evaluated on EUR-LEX, an English corpus introduced in the same paper. We evaluated our English model (Legal BERT), trained on JRC-Acquis and OPOCE, and report in Table \ref{tab:sota_compare} the R-Precision (\textit{RP}), the normalized discounted cumulative gain (nDCG) and the F1-micro score for extracting 5 ID descriptors on the test set of EUR-LEX. Our model obtained a \textit{RP@5} of 81.2\%, a \textit{nDCG@5} of 83.4\% and a micro-F1 of 79.6, outperforming CNN-LWAN, BIGRU-LWAN \cite{mullenbach2018explainable}, BIGRU-LWAN-L2V \cite{chalkidis2019large} and BERT-base. It must be noted that this comparison is not entirely correct because our model was trained on a different corpus which might affect the final results. However, it allows to glimpse the performance of our system in contrast with more modern approaches.

\begin{table}[]
    \centering
    \resizebox{0.48\textwidth}{!}{
        \begin{tabular}{lccc}
            \hline
            \textbf{Model} & \textit{\textbf{RP@5}} & \textit{\textbf{nDCG@5}} & \textit{\textbf{Micro-F1}} \\
            \hline
            CNN-LWAN & 71.6 & 74.6 & 64.2 \\
            BIGRU-LWAN & 76.6 & 79.6 & 69.8 \\
            BIGRU-LWAN-L2V & 77.5 & 80.4 & 71.1 \\
            BERT-base & 79.6 & 82.3 & 73.2 \\ 
            \hline
            \textbf{Legal BERT (ours)} & \textbf{81.2} & \textbf{83.4} & \textbf{79.6} \\ 
            \hline
        \end{tabular}
    }
    \caption{Our Legal BERT model compared with BERT-base and BIGRU-LWAN on EUR-LEX test.}
    \label{tab:sota_compare}
\end{table}

Other extensive document classification experiments with BERT were conducted by \cite{adhikari2019docbert}, without specific consideration to EuroVoc labels. They used a similar approach to ours by introducing a fully-connected layer over the embedding of the first token \texttt{[CLS]}. Furthermore, the paper also presents the results for a knowledge distillation process \cite{hinton2015distilling} from the fine-tuned BERT-large into the previous SOTA \cite{adhikari2019rethinking}, a much smaller network, obtaining better results than BERT-base on the evaluated datasets, but still behind BERT-large.

\subsection{Response Time}

\begin{figure}
    \centering
    \includegraphics[width=0.45\textwidth]{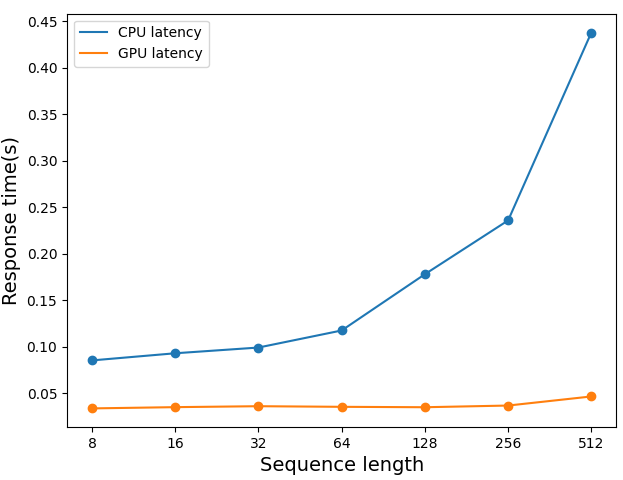}
    \caption{Response time of our API (seconds) for the English language on CPU and GPU.}
    \label{fig:inf_time}
\end{figure}

The response time of the API was tested on a CPU - Intel Xeon Silver 4210 - and on a GPU - Nvidia Quadro RTX 5000. Because the pretrained language models have mostly the same dimension, we made an inference time analysis only for the English variant. Figure \ref{fig:inf_time} depicts the average response time of Legal BERT using various sequence lengths. The response time of the model on GPU is approximately 34 ms on the GPU, with a slight increase to 43 ms when the maximum sequence length of 512 is given as input. However, when the API is run on the CPU the response time increases from 100 ms to 450 ms. 

\section{Programmatic Interface}
\label{sec:prog_interf}

To ease the loading of models and the classification of documents, we created a programmatic interface in Python that can be installed using PyPi with the command \texttt{pip install pyeurovoc}. Once the library is installed, a BERT model is simply created by instantiating the class \texttt{EuroVocBERT} with one of the 22 languages. The class will either download the fine-tuned model from the repository or will use a local cached version of it. Finally, the classification of a document is made by calling the instantiated model with the document text. 

More detailed information about the API and how custom pre-trained BERT models can be fine-tuned on the dataset can be found at the source repository. An example of API usage is presented in Appendix A.

\section{Conclusion and Future Work}
\label{sec:conclusion}

Document classification remains a relevant problem in nowadays society, aiding companies and government institutions to index their large textual database. This paper presented a tool for classifying legal documents with EuroVoc descriptors that use various Transformer-based language models, fine-tuned on the 22 languages that are found in JRC-Acquis and OPOCE. We thoroughly evaluated the models on multiple splits of the data and the results showed that they significantly improve the performance obtained by another similar tool - JEX. The pretrained models were made publicly available and they can be easily used to classify new documents using our API.

One direction for possible future work is to improve the inference speed of the models by either distilling their knowledge in a smaller network \cite{hinton2015distilling} or quantizing their weights \cite{yang2019quantization}. Furthermore, we intend to include our results for legal document classification in language specific NLP benchmarks such as KLEJ for Polish \cite{rybak2020klej}, LiRo for Romanian \cite{dumitrescu2021liro} or EVALITA4ELG for Italian \cite{patti2020evalita4elg}.

\section*{Acknowledgments}
This research was supported by the EC grant no. INEA/CEF/ICT/A2017/1565710 for the Action no. 2017-EU-IA-0136 entitled “Multilingual Resources for CEF.AT in the legal domain” (MARCELL).

\bibliographystyle{acl_natbib}
\bibliography{anthology,ranlp2021}

\appendix

\section{API Code Snippet}
\label{app:API_code}

The following is a code snippet that loads the BERT model for English from the checkpoint repository and classifies a document, given its text.

\begin{lstlisting}[language=Python]

from pyeurovoc import *

model = EuroVocBERT(lang="en")
outputs = model(<document text>)

\end{lstlisting}

The code sniped will return a dictionary of ID descriptors and confidence scores. The number of labels returned by the model for ID descriptors type is controlled by the \texttt{num\_labels}.

\begin{lstlisting}[language=Python]

{
  <ID_label1>: <score1>, 
  <ID_label2>: <score2>,
           ...
}
\end{lstlisting}

\end{document}